\documentclass[runningheads]{llncs}

\usepackage{eccv}

\usepackage{eccvabbrv}
\usepackage{multirow}
\usepackage{graphicx}
\usepackage{booktabs}
\usepackage{caption}

\usepackage[accsupp]{axessibility}  

\usepackage{hyperref}

\usepackage{orcidlink}
\begin{document}

\title{HandSCS: Structural Coordinate Space for Animatable Hand Gaussian Splatting} 

\titlerunning{HandSCS: Structural Coordinate Space for Hand Gaussian Splatting}

\author{Yilan Dong\inst{1}\and
Wenqing Wang\inst{2} \and
Qing Wang\inst{1} \and
Jiahao Yang\inst{1} \and
Haohe Liu\inst{2} \and
Xiatian Zhu\inst{2} \and
Gregory Slabaugh\inst{1} \and
Shanxin Yuan\inst{1}\thanks{Corresponding author.}
}

\authorrunning{Y.~Dong et al.}

\institute{Queen Mary University of London, London, UK \and
University of Surrey, Guildford, UK
}

\maketitle

\begin{abstract}
Photorealistic and animatable hand avatars are essential for applications such as AR/VR, gaming, and telepresence. 
Recent 3D Gaussian Splatting (3DGS) based avatar methods enable real-time rendering of articulated humans, but modeling hands remains challenging due to their compact structure, frequent self-occlusions, and complex finger interactions. 
Existing approaches primarily rely on pose-driven transformations while representing Gaussians in Euclidean space, lacking an explicit structural association with the underlying skeleton, which makes preserving fine-grained hand structures under complex articulations difficult.
In this work, we introduce the Structural Coordinate Space (SCS), a skeleton-relative representation that assigns each Gaussian primitive an explicit structural coordinate with respect to the articulated hand skeleton. 
SCS is constructed using a hybrid static–virtual bone basis together with a distance–rotation structural descriptor that encodes the geometric relationship between Gaussians and bones. 
Based on SCS, we enforce both intra-pose and cross-pose structural consistency by combining per-Gaussian residual embeddings for local appearance modeling with structural correspondence across poses. 
Experiments demonstrate that our approach significantly improves structural consistency and preserves fine geometric details under challenging hand articulations compared with existing 3DGS-based avatar methods.
\end{abstract}

\section{Introduction}
\label{sec:intro}
Rendering photorealistic, animatable hand avatars is essential for applications in AR/VR, gaming, and telepresence. Recent advances in 3D Gaussian Splatting (3DGS)~\cite{3DGS} enable real-time rendering of articulated avatars by representing subjects as collections of 3D Gaussian primitives deformed through skeleton-driven transformations~\cite{gauhuman, 3dgs-avatar, moreau2024human, manus, 3D-PSHR, gaussianhand, gaussianavatars, gaussianbody, splattingavatar}. While this paradigm has shown promising results for human bodies and heads, hands remain particularly challenging. Hand articulation is concentrated in compact structured regions, where visually similar fingers frequently interact and occlude each other. Such tightly coupled, fine-grained motions place stronger demands on maintaining structural consistency across poses in Gaussian-based representations.

\begin{figure*}[tbp]
    \centering
    \includegraphics[width=\linewidth]{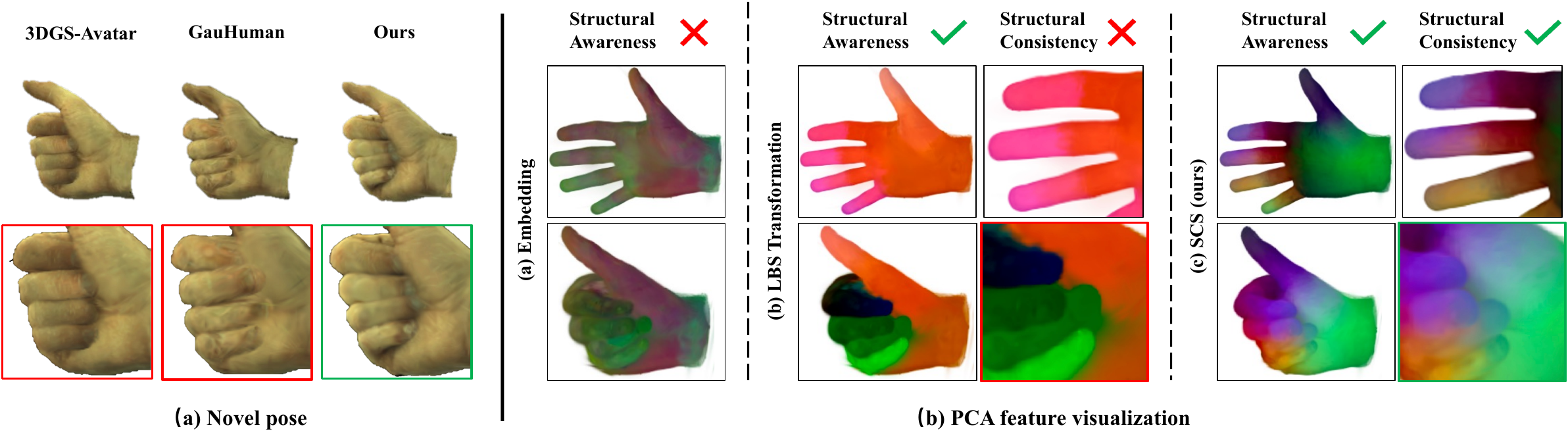}
    \vspace{-1.5em}
    \caption{
    (a) Compared to the 3DGS-Avatar~\cite{3dgs-avatar} and GauHuman~\cite{gauhuman}, our approach can maintain consistency and recover details under novel views and novel poses. 
    (b) Each representation is projected to three principal components mapped to RGB, similar colors indicate similar roles. Color distinction across the hand reflects structural awareness, while consistent coloring of the same region across poses reflects structural consistency. Embedding, nearly uniform, lacking structural awareness. LBS transform, spatially varied but inconsistent across poses. Ours, both structurally aware and pose-consistent.}
    \vspace{-1.5em}
    \label{fig:topology1}
\end{figure*}

Despite the progress of 3DGS-based avatar models, current approaches deform Gaussian primitives primarily through pose-driven transformations, such as global pose conditioning~\cite{3dgs-avatar, gaussianhand, 3D-PSHR} or LBS refinement~\cite{gauhuman, splattingavatar, kocabas2024hugs}. While these mechanisms propagate articulated motion, the Gaussian representation itself is formulated in Euclidean positions and learned attributes, without a disentangled geometric mapping to the underlying skeleton. As illustrated in Fig.~\ref{fig:topology1}, although the LBS transformation matrices provide pose-dependent cues for guiding deformation, these signals vary across poses and do not establish a consistent structural association between a Gaussian and a specific anatomical region. Consequently, maintaining consistent structural behavior under complex articulations becomes challenging. In particular, fine geometric cues, such as fingernail structures and clear separations between adjacent fingers, may degrade in highly articulated configurations where fingers interact closely.

These observations suggest that each Gaussian primitive should carry a structural descriptor relative to the underlying skeleton. However, constructing such a representation is non-trivial. The hand skeleton provides a sparse kinematic scaffold composed of only a small number of joints, whereas 3DGS represents the hand using a large set of spatially distributed Gaussian primitives.
Some methods introduce per-Gaussian latent codes to enhance identity modeling capacity~\cite{moreau2024human, per-gs}. However, as shown in Fig.~\ref{fig:topology1}, the embedding itself does not show any semantic region-related distribution. Then, without the underlying structure guidance, a Gaussian can drift and lie in different semantic roles under different poses, while the embedding is fixed across these semantic roles, leading to smoothed-out representations and loss of fine-grained details.

To address this challenge, we introduce the Structural Coordinate Space (SCS), which consists of two components: a hybrid static–virtual bone basis that defines a structural coordinate system for Gaussian primitives, and a distance–rotation transformation that describes the geometric relationship between a Gaussian and the bone basis.
We construct the bone basis by combining static bones in posed space derived from the MANO~\cite{mano} kinematic topology with additional cross-finger connections and removing redundant connections. To further enrich the structural representation, we introduce pose-related virtual bones in canonical space that provide complementary structural cues in highly articulated regions. While static bones provide stable global anatomical coverage, virtual bones expand the basis to better capture pose-dependent structural variations.
The second component is a \textit{structural coordinate descriptor}, the transformation that maps each Gaussian's physical position into its coordinate in SCS, defined through angular and radial terms for each bone in the basis. Together, the basis and descriptor assign every Gaussian an explicit, interpretable structural coordinate within the hand anatomy.

With this representation, each Gaussian can be associated with a structural attribute relative to the articulated skeleton. We then promote consistency from two aspects:
(1) Within a given pose, SCS assigns each Gaussian a distinct structural identity. Conditioned on this structural reference, we introduce per-Gaussian embeddings~\cite{moreau2024human, per-gs} to capture local geometric and appearance variations. These embeddings parameterize residual offsets for standard Gaussian attributes, enabling flexible detail modeling while preserving the structural identity provided by SCS.
(2) Across poses, Gaussians with similar SCS structural coordinates exhibit consistent behavior under similar articulation. 
To enforce this, we establish cross-pose correspondences by matching Gaussians through their SCS structural coordinates, and penalize attribute divergence between corresponding Gaussians, modulated by pose similarity.

Our contributions are summarized as follows:
\begin{enumerate}
    \item We introduce the Structural Coordinate Space (SCS), a skeleton-relative representation that assigns each Gaussian primitive an explicit structural coordinate with respect to the articulated hand skeleton.
    \item Built upon SCS, we design a structure-aware learning scheme that enforces both intra-pose and cross-pose consistency through per-Gaussian residual embeddings and structural correspondence across poses.
    \item Experiments on the InterHand2.6M dataset demonstrate that HandSCS achieves state-of-the-art in hand avatar animation, effectively preserves fine-grained geometric details under complex hand articulations while maintaining real-time rendering speed.
\end{enumerate}

\section{Related Work}
\label{sec:related}

\subsection{Mesh based Hand Avatars}
Parametric meshes are widely used for hand avatar modeling~\cite{s2hand, dhm, amvur, nimble, html, mano, harp, xhand, handy, moon2024authentic, fine} due to their explicit geometry and compatibility with animation pipelines.
Early works~\cite{s2hand, dhm} estimate colored meshes, but lack fine-grained appearance.
Recent methods enhance realism by incorporating appearance modeling: AMVUR~\cite{amvur} introduces attention-based vertices and occlusion-aware texture regression; NIMBLE~\cite{nimble} builds a non-rigid model with bones, muscles, and physically-based skin; HTML~\cite{html} extends MANO~\cite{mano} with a PCA-based texture model; and HARP~\cite{harp} adds explicit albedo and normal maps.
However, these methods often rely on dense surface capture, non-rigid tracking, or subject-specific templates, increasing cost and limiting scalability.

\subsection{Neural Rendering for Hand Avatars}
Neural implicit representations~\cite{nerf, jiang2022selfrecon, mihajlovic2021leap} have been widely adopted for animatable hand avatar. Many methods~\cite{lisa, handavatar, handnerf, livehand, phrit, ohta, bitt} render hands via inverse LBS and volume rendering of predicted color and density fields.
LISA~\cite{lisa} predicts per-bone color and signed distance fields; HandAvatar~\cite{handavatar} adds explicit albedo and illumination modeling; HandNeRF~\cite{handnerf} introduces hand-specific deformation fields.
While producing high-quality results, these methods suffer from slow rendering due to dense ray sampling. 
To improve efficiency, mesh-guided acceleration has been explored~\cite{livehand, phrit}, though PHRIT~\cite{phrit} remains slow (~10 FPS) and LiveHand~\cite{livehand} sacrifices realism for speed.
OHTA~\cite{ohta} and BiTT~\cite{bitt} reconstruct relightable, personalized appearance from a single image, but struggle to maintain consistency across poses and viewpoints.

\subsection{3DGS based Hand Avatars}
3D Gaussian Splatting (3DGS~\cite{3DGS}) provides a new paradigm for real-time scene reconstruction, and has been adept at representing hand avatars by modeling geometry and appearance through continuous and differentiable Gaussian primitives ~\cite{3D-PSHR, gaussianhand, manus, InterHandGS, jghand}.
3D-PSHR~\cite{3D-PSHR} incorporates albedo and shading into the Gaussian representation, but lacks the ability to model pose-dependent appearance.
GaussianHand~\cite{gaussianhand} improves rendering quality by learning Gaussian offsets from pose-dependent and pose-independent properties with a UNet-based appearance decoder.
However, existing approaches do not explicitly model the underlying hand structure, limiting their ability to maintain stable geometry and appearance under challenging cases involving large articulations.

\subsection{3DGS based Human Avatars}
Recent works have explored animatable human avatars using 3D Gaussian Splatting~\cite{drivable, gauhuman, splattingavatar, 3dgs-avatar, kocabas2024hugs, hifi4g, ash, gps, kwon2024generalizable, gomavatar, moreau2024human, li2024animatable}.  
Some methods~\cite{gauhuman, splattingavatar, kocabas2024hugs, gomavatar} reconstruct avatars from monocular or multi-view videos, but do not explicitly model pose-dependent appearance.  
To address this, several approaches~\cite{3dgs-avatar, moreau2024human} learn Gaussian property offsets via a MLP, conditioned on the pose along with either positions~\cite{3dgs-avatar} or per-Gaussian latent codes~\cite{moreau2024human}.  
These methods have demonstrated promising results for clothed human bodies, which often involve challenges such as loose garments and topology variation.
However, hand modeling introduces different difficulties, including high articulation, self-occlusion, and inter-part contact, highlighting the need for structure-aware modeling, which prior Gaussian-splatting avatars have not explored.

\section{Method}
\begin{figure*}[htbp]
    \centering
    \vspace{-1em}
    \includegraphics[width=\textwidth]{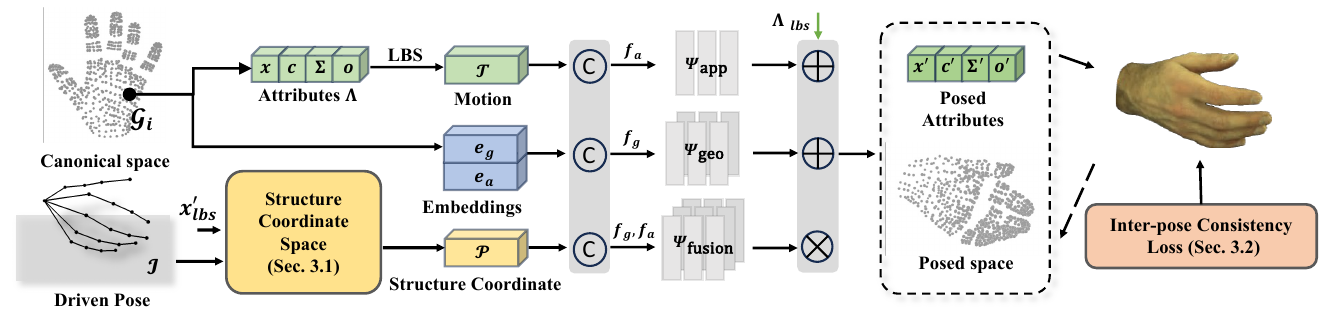}
\caption{
\textbf{Overview of HandSCS}.
Each 3D Gaussian is modeled in canonical space and reparameterized in SCS. The structural coordinate $\mathcal{P}$ is computed from the LBS-posed position $x'_{\textit{lbs}}$ and the skeleton joints via angular-radial descriptors over the hybrid static-virtual bone basis. Together with per-Gaussian geometry embedding $e_g$ and appearance embedding $e_a$, the structural coordinate determines non-rigid attribute offsets. Cross-pose consistency is enforced by matching Gaussians through their structural coordinates in SCS and penalizing attribute divergence.
}
    \vspace{-2em}
    \label{fig:pipeline}
\end{figure*}

Fig.~\ref{fig:pipeline} shows the framework of HandSCS. We model a 3D hand in canonical space, where each MANO mesh vertex is initialized as a 3D Gaussian with attributes $\mathbf{\Lambda}=\{\mathbf{x},  \mathbf{c}, \mathbf{\Sigma}, \mathbf{o}\}$ representing position, color, covariance, and opacity. The core of our approach is the Structural Coordinate Space (SCS), which reparameterizes each Gaussian by its geometric relationship to the hand skeleton (Sec.~\ref{sec:3.1}). We define SCS through its coordinate basis, the set of bones forming the reference frame (Sec.~\ref{sec:3.1.1}), and a coordinate descriptor, the transformation that maps each Gaussian into this space (Sec.~\ref{sec:3.1.2}). Non-rigid deformation is formulated within SCS, with each Gaussian's attribute offsets determined by its structural coordinate and per-Gaussian embeddings (Sec.~\ref{sec:3.1.3}). To promote consistency across poses, we introduce an inter-pose consistency loss that establishes correspondences through structural coordinates in SCS (Sec.~\ref{sec:3.2}).

\subsection{Structural Coordinate Space}
\label{sec:3.1}

\subsubsection{Structural Coordinate Basis}
\label{sec:3.1.1}
We construct the coordinate basis from a unified set of static and virtual bone vectors.
The static bones capture the unified physical hand structure and are deformed through the standard LBS in MANO model, while preserving a consistent canonical skeletal topology across poses. In contrast, the virtual bones adaptively model pose-dependent structural variations.
Together, they form a flexible structural coordinate basis.

\begin{figure*}[tbp]
    \centering
    \includegraphics[width=\linewidth]{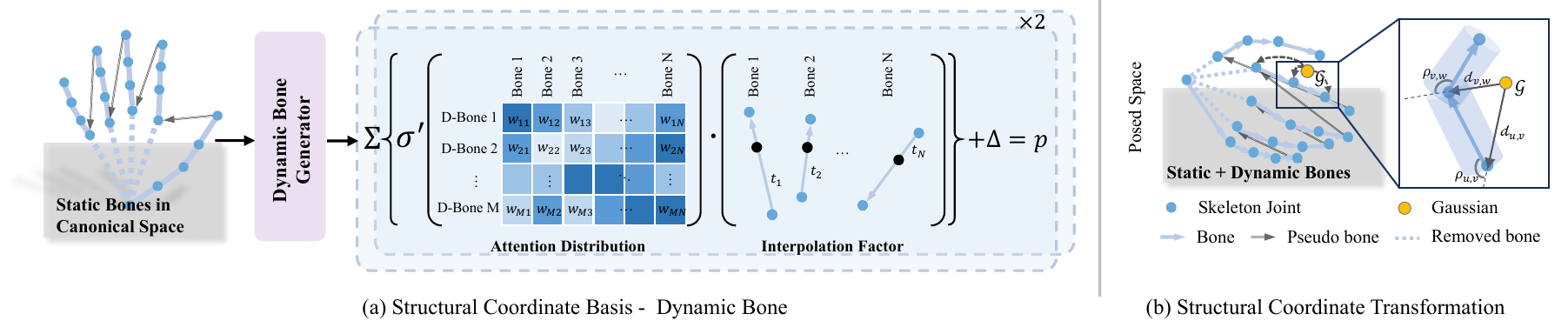}
    \caption{Illustration of \textbf{Structural Coordinate Space} (SCS). The left part shows the extended kinematic topology $\mathcal{E}$ based on MANO model, with added cross-finger pseudo-bones and removal of redundant connections. The right part illustrates how each Gaussian is encoded with its cosine of angle $\rho_{u,v}$ and distance $d_{u,v}$ relative to each bone in the posed space.} 
    \vspace{-1em}
    \label{fig:topology}
\end{figure*}

\noindent\textbf{Static Bone Basis.}
The static part defines the physical hand bones derived from the MANO kinematic model~\cite{mano}, augmented with cross-finger connections to capture anatomical relationships between adjacent fingers. 
To maintain a compact topology, we remove redundant wrist-to-finger edges for all fingers except the thumb, as illustrated in Fig.~\ref{fig:topology}.
The resulting static bone topology is formulated as,
\begin{equation}
\mathcal{E}_{\text{s}} = 
\big( \mathcal{E}_{\text{MANO}} \setminus \mathcal{E}_{\text{removed}} \big)
\cup \mathcal{E}_{\text{cross}},
\end{equation}
Given the pose parameters $\boldsymbol{\theta}$ and shape parameters $\boldsymbol{\beta}$, the MANO model computes the joint positions in observed space $\mathcal{J}_o = \{\boldsymbol{j}_o^i\}_{i=0}^{K-1}
$,
where $\mathcal{J}_o \in \mathbb{R}^{K \times 3}$ and $K$ is the number of joints. 
Using the defined topology $\mathcal{E}_{\text{static}} = \{ (u, v) \mid v \in \mathcal{J} \}$, 
where $u$ denotes the parent joint of $v$. 
The set of static bones is formulated as,
\begin{equation}
\mathcal{B}_{\text{s}} = 
\{ \mathbf{b}_{u,v}=(\boldsymbol{j}_o^u, \boldsymbol{j}_o^v) \mid (u,v) \in \mathcal{E}_{\text{s}} \},
\end{equation}
Here, “static” refers to the fixed skeletal topology.

\noindent{\bf{Virtual Bone Basis.}}
The predefined static bone alone cannot fully capture the structural variations induced by highly articulated or interacting poses.
To address this limitation, we introduce virtual bones, constructed by referencing the static bones and interpolating their endpoints along bone segments.

Formally, given the static bone topology $\mathcal{E}_{\text{static}}$ and the canonical joint locations 
$\mathcal{J} = \{\boldsymbol{j}^i\}_{i=0}^{K-1}$, 
we can obtain the set of canonical static bones as 
$\mathcal{B}_{\text{s}}^{c} = \{ (\boldsymbol{j}^u, \boldsymbol{j}^v) \mid (u,v) \in \mathcal{E}_{\text{static}} \}$.
The virtual bone $\mathcal{B}_{\text{d}}$ is represented by two endpoints $(p, q)$.
For virtual bone $i$, the model (i) predicts an attention distribution $w$ over all static bones to identify relevant reference regions, and (ii) learns interpolation factors $t$ and offsets $\Delta$ to determine precise start and end positions.
The endpoints are computed as,
\begin{equation}
\begin{aligned}
p &= \sum_{b}^{\mathcal{B}} \tilde{w}^s_{b} \big[(1 - t_b^s)\mathbf{j}^u + t_b^s \mathbf{j}^v\big] + \Delta p_b, \\
q &= \sum_{b}^{\mathcal{B}} \tilde{w}^e_{b} \big[(1 - t_b^e)\mathbf{j}^u + t_b^e \mathbf{j}^v\big] + \Delta q_b,
\end{aligned}
\end{equation}
where
$w^s_{b}$ and $w^e_{b}$ are the attention weights over canonical static bones, and $t^s, t^e \in [0,1]$ denote the interpolation factors along each bone, and $\Delta \mathbf{p}$ and $\Delta \mathbf{q}$ are pose-dependent offsets for fine-grained adjustments.
The final structural coordinate basis is obtained as $\mathcal{B} = \mathcal{B}_s \cup \mathcal{B}_d$ and the topology is $\mathcal{E} = \mathcal{E}_s \cup \mathcal{E}_d$.

To encourage smooth transitions between adjacent bones, we apply a smoothing kernel to the attention distribution. 
Given the attention logits \(\mathcal{W}=\{w_b|b\in\mathcal{E}_{\text{static}}\} \), the smoothed weights are computed as,
\begin{equation}
\tilde{\mathcal{W}} \;=\; \frac{K\,\sigma(\mathcal{W})}{\big\|K\,\sigma(\mathcal{W})\big\|_1}
\end{equation}
where \(\sigma(\cdot)\) denotes the softmax operation, and \(K \in \mathbb{R}^{B \times B}\) is a fixed topology-aware smoothing kernel defined over the static hand bone, assigning a weight of \(0.5\) to each bone itself, \(0.25\) to its directly connected neighbors and \(0\) to all other bones.

In practice, we generate $M$ virtual bones using a lightweight MLP as the virtual bone generator.
The network takes the pose parameters $\boldsymbol{\theta}$ and canonical joint positions $\mathcal{J}$ as input, and outputs the parameters ${w^s, w^e, t^s, t^e, \Delta \mathbf{p}, \Delta \mathbf{q}}$.
The resulting virtual bones are then combined with the static ones to form an enriched structural prior, providing a more flexible coordinate basis.
The static bones define a stable anatomical reference frame; the virtual bones expand this basis to capture pose-dependent structural variations, enabling the coordinate space to adapt to the current articulation.

\subsubsection{Structural Coordinate Descriptor}
\label{sec:3.1.2}
The coordinate descriptor maps each Gaussian's physical position into its structural coordinate in SCS, encoding its geometric relationship to each bone in the basis.

\noindent\textbf{Per-bone structural relations.}
Let $\mathbf{x}'_{\text{lbs}}$ denote the LBS-posed center of a Gaussian.
The structural relation between $\mathbf{x}'_{\text{lbs}}$ and the coordinate basis $\mathcal{B}$ is decomposed into 
an \emph{angular} term and a \emph{radial} term.
The angular term measures how the Gaussian is oriented with respect to the bone. For each bone $\textbf{b}_{u,v} = (j^u, j^v)$, 
we define the Gaussian-to-basis vector 
$\mathbf{r}_{u,v} = \mathbf{j}_u - \mathbf{x}'_{\text{lbs}}$.
We use the cosine similarity between $\mathbf{r}_{u,v}$ and $\mathbf{b}_{u,v}$,
\begin{equation}
\rho_{u,v} =
\frac{\mathbf{r}_{u,v} \cdot \mathbf{b}_{u,v}}
{\|\mathbf{r}_{u,v}\|_2 \, \|\mathbf{b}_{u,v}\|_2},
\quad (u,v)\in\mathcal{E},
\end{equation}
where $\| \cdot \|_2$ denotes L2 norm.
The radial term measures the spatial affinity between the Gaussian and the bone. For static bones in posed space, we apply an exponential decay,
\begin{equation}
d_{u,v} =
\exp\!\left(-\frac{\|\mathbf{r}_{u,v}\|_2^2}{\tau^2}\right),
\quad (u,v)\in\mathcal{E}_{\text{s}},
\end{equation}
where $\tau$ is the decay rate, set to $0.08$, restricting each bone's influence to its local neighborhood.
For virtual bones in canonical space, we use the raw distance,
\begin{equation}
d_{u,v} = \|\mathbf{r}_{u,v}\|_2, \quad (u,v)\in\mathcal{E}_{\text{d}},
\end{equation}
which preserves a larger spatial dynamic range, providing complementary structural cues to the bounded static bone descriptors.

\noindent\textbf{Structural coordinates.}
By aggregating these relations across all bones,
the structural coordinate descriptor for each Gaussian is defined as,
\begin{equation}
\mathcal{P}=[d_{u,v}\cdot\rho_{u,v}]_{(u,v)\in\mathcal{E}}.
\end{equation}

where $[\,\cdot\,]$ denotes concatenation. Each dimension of $\mathcal{P} \in \mathbb{R}^{|\mathcal{E}|}$ encodes the Gaussian's structural relationship to one bone in the basis. The resulting structural coordinate is spatially smooth, as nearby Gaussians obtain similar coordinates, yet structurally discriminative: Gaussians from different structural regions maintain distinct coordinates even when spatially close, such as during self-contact between fingers.

\subsubsection{Intra-pose Consistency in SCS.}
\label{sec:3.1.3}
In SCS, each Gaussian's identity consists of two complementary aspects: its structural coordinate $\mathcal{P}$, which encodes its geometric relationship to the skeleton, and per-Gaussian learnable embeddings that capture fine-grained variations specific to each Gaussian. We assign each Gaussian a geometry embedding $\mathbf{e}_g \in \mathbb{R}^{d}$ and an appearance embedding $\mathbf{e}_a \in \mathbb{R}^{d}$, jointly optimized through differentiable rendering. While $\mathcal{P}$ captures the Gaussian's structural role within the hand anatomy, the embeddings model subject-specific surface details that vary across individual Gaussians at the same structural position.

Non-rigid deformation is determined jointly by the structural coordinate $\mathcal{P}$, per-Gaussian embeddings, and the LBS transformation matrix $\mathcal{T}$.
Geometry and appearance offsets are decoded from these components through MLPs. The final attributes combine canonical values with the decoded offsets,
\begin{equation}
\begin{aligned}
\mathbf{x}' &= \mathbf{x} + \Delta \mathbf{x} + \Delta \mathbf{x}^{f}, \\
c' &= c + \Delta c + \Delta c^{f}, \\
\boldsymbol{\Sigma}' &= (\Delta \boldsymbol{\Sigma}^{f} \Delta \boldsymbol{\Sigma}) \,\boldsymbol{\Sigma}\, (\Delta \boldsymbol{\Sigma} \Delta \boldsymbol{\Sigma}^{f}).
\end{aligned}
\end{equation}
where $\Delta \mathbf{x}, \Delta \boldsymbol{\Sigma}, \Delta c$ are decoded from geometry or appearance representations and $\Delta \mathbf{x}^{f}, \Delta \boldsymbol{\Sigma}^{f}, \Delta c^{f}$ from their fusion (details in supplementary material).

Each component contributes a distinct aspect: $\mathcal{T}$ captures rigid bone-driven motion, the embeddings capture per-Gaussian identity, and $\mathcal{P}$ encodes each Gaussian's position within the hand's skeletal structure. This enables the deformation to reflect structural role: Gaussians near joints, at fingertips, or in contact regions each receive structurally appropriate offsets determined by their position in SCS.

\subsection{Inter-Pose Consistency in SCS.}
\label{sec:3.2}
The structural coordinate assigned to each Gaussian encodes its relationship to the skeleton rather than its absolute spatial position, making it a natural basis for establishing correspondences across different poses. We leverage this property to enforce cross-pose consistency: Gaussians occupying the same structural role should exhibit coherent attributes across similar articulations.

At training iteration $t$, each Gaussian $\mathcal{G}_{t}^{i}$ is associated with pose parameter $\boldsymbol{\theta}_t$, structural coordinate $\mathcal{P}_{t}^{i}$ in SCS, and attributes $\mathbf{M}_t^{i} = (\mathbf{x}_t^i, \boldsymbol{\Sigma}_t^i, \mathbf{c}_t^i, \mathbf{e}_{g,t}^i, \mathbf{e}_{a,t}^i)$. To modulate the loss, we first compute the pose similarity as weighting factor,
\begin{equation}
\omega_{t, t-1} = \exp\left(-\frac{\|\Phi_{\theta}(\boldsymbol{\theta}_t) - \Phi_{\theta}(\boldsymbol{\theta}_{t-1})\|_2^2}{2\delta^2}\right),
\end{equation}
where $\Phi_{\theta}$ is a pose encoder and $\delta$ is set to 1.3.

Correspondences between the current and previous frames are established through structural coordinates. For each Gaussian $\mathcal{G}_{t}^{i}$, we find its structurally nearest counterpart in the previous frame,
\begin{equation}
\pi(i) = \arg\min_{j=1}^{N_{t-1}} \ \|\mathcal{P}_{t}^{i} - \mathcal{P}_{t-1}^{j}\|_2^2,
\end{equation}
The consistency loss penalizes attribute divergence between structurally corresponding Gaussians,
\begin{equation}
\mathcal{L}_{\text{con}} = \omega_{t,t-1}\sum_{i=1}^{N_t} \| \mathbf{M}_{t}^i- \mathbf{M}_{t-1}^{\pi(i)} \|_2^2,
\end{equation}
where $\mathbf{M}_{t-1}^{\pi(i)}$ denotes attributes from last iteration. For initialization, $\mathbf{M}_{0}^i = \mathbf{M}_{1}^i$.

By including $\mathbf{e}_g, \mathbf{e}_a$,
the consistency loss also regularizes the non-rigid deformation to be coherent across structurally corresponding Gaussians.
Also, when $\boldsymbol{\theta}_t = \boldsymbol{\theta}_{t-1}$, the loss naturally enforces cross-view alignment for the same pose, improving robustness under self-occlusion and viewpoint variation.

\subsection{Training Objective Function}
Following other 3DGS-based methods \cite{manus, 3D-PSHR, 3dgs-avatar, gauhuman}, we apply a standard RGB loss, mask loss, SSIM loss, and LPIPS loss to supervise the 3D model,
\begin{equation}
    \mathcal{L}_{\text{base}} = \mathcal{L}_{\text{rgb}} + \lambda_1 \mathcal{L}_{\text{mask}} + \lambda_2 \mathcal{L}_{\text{SSIM}} + \lambda_3 \mathcal{L}_{\text{LPIPS}}, \label{equ: standard loss}
\end{equation}
where $\lambda$ are loss weights. Empirically, we set $\lambda_1 = 0.1$, $\lambda_2 = \lambda_3 = 0.01$. Additionally, to prevent the adaptive attribute offsets from producing excessively large deformations, we apply an offset regularization loss,
\begin{equation}
    \mathcal{L}_{\text{reg}} = \| \Delta \mathbf{x} \|_2 + \| \Delta \boldsymbol{\Sigma} - \mathbf{I} \|_F + \| \Delta c \|_2,
\end{equation}
where $\| \cdot \|_F$ denotes the Frobenius norm. This term penalizes the position and color offset magnitudes while encouraging the covariance offset to remain close to the identity matrix, preventing unstable shape distortions. 
The overall loss is,
\begin{equation}
\mathcal{L} = \mathcal{L}_{\text{base}} + \lambda_{\text{con}}\mathcal{L}_{\text{con}} + \lambda_{\text{reg}}\mathcal{L}_{\text{reg}},
\end{equation}
where $\lambda_{\text{con}}$ is set to 0.01, and $\lambda_{\text{reg}}$ is set to 0.1.

\section{Experiments}

\begin{figure*}[tbp]
    \centering
    \includegraphics[width=\textwidth]{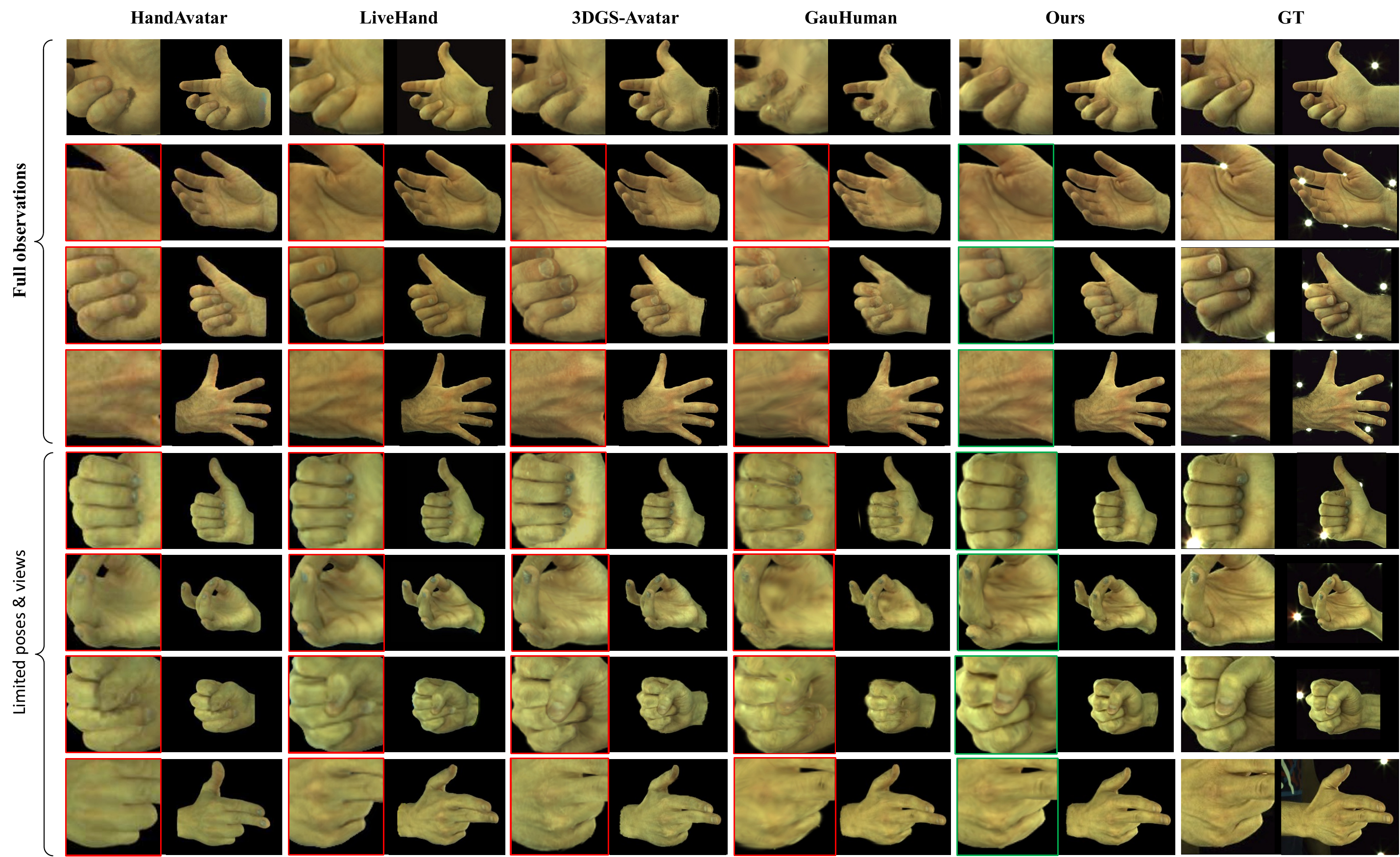}
    \caption{{\bf Qualitative Comparison of Novel Pose Synthesis.} We compare results on InterHand2.6M \cite{interhand} across five subjects using HandAvatar \cite{handavatar}, LiveHand \cite{livehand}, GauHuman \cite{gauhuman}, and our method. HandSCS produces sharper structures, cleaner boundaries, and fewer artifacts under challenging articulations.
    }
    \label{fig:vis_full}
    \vspace{-1em}
\end{figure*}

\subsection{Datasets and Metrics}
We conduct experiments on the widely used InterHand2.6M \cite{interhand} dataset at 5 FPS, which provides large-scale multi-view sequences of diverse hand poses. 
To fairly assess our method, we select right-hand sequences from five subjects as shown in Tab.~\ref{tab:model_comparison}, where the last two subjects are sparse-view and sparse-pose tasks. For novel pose rendering, we allocate the last 50 poses for testing and use the remaining poses for training. For novel view synthesis, 10 views are selected for training, 15 views are used as testing set. Detailed specifications of the selected camera views are provided in the supplementary material.

We evaluate Peak Signal-to-Noise Ratio (PSNR), Structural Similarity Index Measure (SSIM)~\cite{ssim} and Learned Perceptual Image Patch Similarity (LPIPS)~\cite{lpips} as metrics for the rendering results on full image. We use FPS to compare the rendering speed. Rendering speeds are evaluated on a single NVIDIA A100 GPU.

\begin{table*}[h!]
    \vspace{-1em}
    \caption{\textbf{Novel pose animation on InterHand2.6M \cite{interhand}.}
    HandSCS achieves the best overall performance. Bold numbers indicate the best results.}
    \scriptsize
    \centering
    \resizebox{\linewidth}{!}{%
        \begin{tabular}{lccccccccccccccc}
        \toprule[1pt]
        \multirow{2}{*}{Method}& \multicolumn{3}{c}{test/Capture0}& \multicolumn{3}{c}{test/Capture1} & \multicolumn{3}{c|}{val/Capture0} & \multicolumn{3}{c}{train/Capture5} & \multicolumn{3}{c}{train/Capture6} 
        \\
        \cmidrule(lr){2-4} \cmidrule(lr){5-7} \cmidrule(lr){8-10} \cmidrule(lr){11-13} \cmidrule(lr){14-16}
         & PSNR $\uparrow$ & SSIM $\uparrow$ & LPIPS $\downarrow$  & PSNR $\uparrow$ & SSIM $\uparrow$ & LPIPS $\downarrow$ & PSNR $\uparrow$ & SSIM $\uparrow$ & LPIPS $\downarrow$ & PSNR $\uparrow$ & SSIM $\uparrow$ & LPIPS $\downarrow$ & PSNR $\uparrow$ & SSIM $\uparrow$ & LPIPS $\downarrow$ \\
        
        \midrule
        
        \multicolumn{1}{l|}{LiveHand~\cite{livehand}}  &  29.17 & 0.949 & \multicolumn{1}{c|}{0.0528} & 26.85 & 0.776 & \multicolumn{1}{c|}{0.0619} & 29.33 & 0.893 & \multicolumn{1}{c|}{0.0606} & 31.35 & \bf{0.975} & \multicolumn{1}{c|}{0.0238} & 30.67 & \bf{0.969} & \multicolumn{1}{c}{0.0274}  \\
        \multicolumn{1}{l|}{HandAvatar~\cite{handavatar}}  & 31.36  &  0.959 & \multicolumn{1}{c|}{0.0453} & 30.37 & 0.961 & \multicolumn{1}{c|}{0.0448} & 30.35 & 0.954 & \multicolumn{1}{c}{0.0499} & 30.36 & 0.966 & \multicolumn{1}{c}{0.0315} & 29.33 & 0.958 & \multicolumn{1}{c}{0.0390} \\
        \midrule
        \multicolumn{1}{l|}{3dgs-avatar~\cite{3dgs-avatar}} & 31.35 & 0.953 & \multicolumn{1}{c|}{0.0477} &  31.07 & 0.957 & \multicolumn{1}{c|}{0.0475} & 31.12 & 0.950 & \multicolumn{1}{c|}{0.0498} &  30.80 & 0.965 & \multicolumn{1}{c|}{0.0304} & 29.59 & 0.956 & \multicolumn{1}{c}{0.0356} \\
        \multicolumn{1}{l|}{GauHuman~\cite{gauhuman}} & 31.71 & 0.958 & \multicolumn{1}{c|}{0.0424} & 31.72 & 0.962 & \multicolumn{1}{c|}{0.0445} & 31.15 & 0.955 & \multicolumn{1}{c|}{0.0488} & 
        31.06 & 0.965 & \multicolumn{1}{c|}{0.0304} & 30.53 & 0.960 & \multicolumn{1}{c}{0.0354} \\
        \midrule
        \multicolumn{1}{l|}{HandSCS} & \bf{32.91} & \bf{0.966} & \multicolumn{1}{c|}{\bf{0.0291}} &\bf{32.99} & \bf{0.970} & \multicolumn{1}{c|}{\bf{0.0282}} & \bf{32.40} & \bf{0.963} & \multicolumn{1}{c|}{\bf{0.0328}} &\bf{32.09} & 0.972 & \multicolumn{1}{c|}{\bf{0.0228}} & \bf{31.31} & 0.966 & \multicolumn{1}{c}{\bf{0.0273}} \\
        
        \bottomrule[1pt]
    \end{tabular}%
    }
    \label{tab:model_comparison}
\end{table*}

\begin{table*}[h!]
    \vspace{-1em}
    \caption{\textbf{Sparse-view novel view synthesis on InterHand2.6M \cite{interhand}.}
    Models are trained on 10 views and evaluated on 15 held-out views. 
    While LiveHand \cite{livehand} attains slightly higher SSIM due to smoother NeRF interpolation under sparse viewpoints, 
    HandSCS produces more consistent structures, clearer boundaries, and fewer artifacts under challenging viewing angles. 
    Bold numbers indicate the best results.}
    \scriptsize
    \centering
    \resizebox{\linewidth}{!}{%
        \begin{tabular}{lccccccccc}
        \toprule[1pt]
        \multirow{2}{*}{Method}& \multicolumn{3}{c}{test/Capture0}& \multicolumn{3}{c}{test/Capture1} & \multicolumn{3}{c}{val/Capture0} 
        \\
        \cmidrule(lr){2-4} \cmidrule(lr){5-7} \cmidrule(lr){8-10} 
         & PSNR $\uparrow$ & SSIM $\uparrow$ & LPIPS $\downarrow$  & PSNR $\uparrow$ & SSIM $\uparrow$ & LPIPS $\downarrow$ & PSNR $\uparrow$ & SSIM $\uparrow$ & LPIPS $\downarrow$ \\
    
        \midrule
        \multicolumn{1}{l|}{LiveHand~\cite{livehand}}  &  31.95 & \textbf{0.984} & \multicolumn{1}{c|}{0.0335} & 31.03 & \textbf{0.979}  & \multicolumn{1}{c|}{0.0367} & 30.21 & 0.969 & \multicolumn{1}{c}{0.0449} \\
        \multicolumn{1}{l|}{HandAvatar~\cite{handavatar}}  &  32.06 & 0.961  & \multicolumn{1}{c|}{0.0440} & 31.62 & 0.963 & \multicolumn{1}{c|}{0.0440} & 32.74 & 0.963 & \multicolumn{1}{c}{0.0432} \\
        \midrule
        \multicolumn{1}{l|}{3dgs-avatar~\cite{3dgs-avatar}} & 31.77 & 0.953 & \multicolumn{1}{c|}{0.0512} & 31.56 & 0.956 & \multicolumn{1}{c|}{0.0479} & 32.20 & 0.953 & \multicolumn{1}{c}{0.0498} \\
        
        \multicolumn{1}{l|}{GauHuman~\cite{gauhuman}} & 32.10 & 0.958 & \multicolumn{1}{c|}{0.0446} & 31.37 & 0.958 & \multicolumn{1}{c|}{0.0462} & 31.73 & 0.955 & \multicolumn{1}{c}{0.0479} \\
        \midrule
        \multicolumn{1}{l|}{HandSCS} & \bf{35.18} & 0.973 & \multicolumn{1}{c|}{\bf{0.0297}} &\bf{35.41} & 0.975 & \multicolumn{1}{c|}{\bf{0.0283}} & \bf{35.74} & \bf{0.972} & \multicolumn{1}{c}{\bf{0.0300}} \\
        \bottomrule[1pt]
    \end{tabular}%
    }
    \label{tab:model_comparison_view}
    \vspace{-2em}
\end{table*}

\subsection{Implementation details}
All three offset decoders, geometry, appearance, and fusion share the same structure. We adopt the same architecture, consisting of a fully connected multilayer perceptron with 2 hidden layers, each containing 256 neurons, a learnable gate (initialized to zero) is included to avoid very large outputs at the beginning of training. 
For Gaussian initialization, we use the MANO model with 778 vertices as the initial state. The color and opacity of each Gaussian are randomly initialized. More details are provided in our supplementary materials and code\footnote{Code: \url{https://github.com/yilan120/HandSCS}}.

\subsection{Baselines and Result Comparison}
\noindent\textbf{Baselines.} To evaluate the performance of HandSCS, we conduct comparisons with state-of-the-art NeRF-based method HandAvatar~\cite{handavatar} and LiveHand~\cite{livehand} and Gaussian Splatting-based methods 3DGS-Avatar~\cite{3dgs-avatar} and GauHuman~\cite{gauhuman}.
To ensure fairness, we include only open-source methods and use their official implementations with default configurations. 

\noindent\textbf{Novel Pose Synthesis.}
Tab.~\ref{tab:model_comparison} summarizes the novel pose results across five subjects. HandSCS achieves the best performance on all metrics, with average gains of +1.24 dB PSNR, +0.008 SSIM, and a 33.4\% relative LPIPS reduction over the strongest baseline. These improvements are comparable to or larger than those reported in recent works~\cite{livehand, handavatar, 3dgs-avatar, gauhuman}.
Qualitative examples in Fig.~\ref{fig:vis_full} show that HandSCS produces sharper details, clearer boundaries, and more stable appearance in challenging articulations. In contrast, existing methods often struggle with fine structures such as fingernails and skin textures or exhibit noticeable artifacts due to limited structural modeling.

\begin{figure}[h!]
\vspace{-1em}
\begin{minipage}{0.48\linewidth}
\centering
\includegraphics[width=\linewidth]{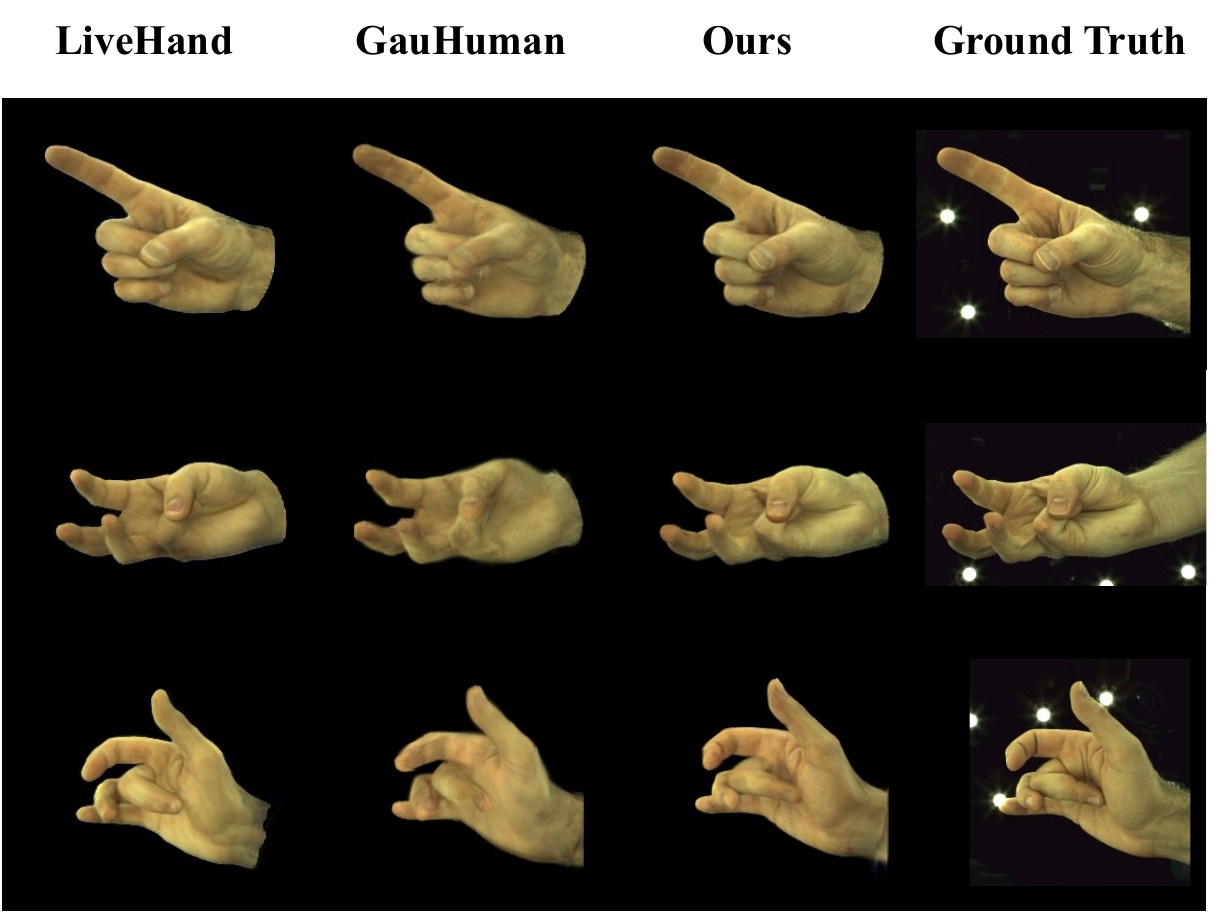}
\caption{{\bf Visual comparison of Novel View Synthesis.}}
\label{fig:novel-view_table}
\end{minipage}
\hfill
\centering
\begin{minipage}{0.48\linewidth}
\centering
\resizebox{\linewidth}{!}{
\begin{tabular}{lccc}
    \toprule
    \textbf{Models}
      & \textbf{PSNR} $\uparrow$ & \textbf{SSIM} $\uparrow$ & \textbf{LPIPS} $\downarrow$ \\
    \midrule
    (a) baseline & 31.71 & 0.958 & 0.0428 \\
    (b) Embedding+(a) & 32.23 & 0.961 & 0.0337 \\
    \midrule
    \hspace{1.7em}$\mathcal{L}_{\text{reg}}$ + (b) & 32.32 & 0.963 & 0.0339 \\
    (c) Inter-pose+(b) & 32.42 & 0.964 & 0.0335 \\
    (d) SCS(Intra-pose)+(b) & 32.78 & 0.966 & 0.0301 \\  
    \midrule
    Full model & \textbf{32.91} & \textbf{0.966} & \textbf{0.0291}  \\
    \bottomrule
\end{tabular}}
\captionof{table}{{\textbf{Ablation study on multi-component structural guidance.}}
    (b) introduces disentangled embeddings, and (c)–(d) further incorporate inter-pose and intra-pose (SCS) structural guidance.
    All components contribute measurable gains, and the full model achieves the best overall performance.
    }
\label{tab:component_table}
\end{minipage}
\vspace{-4em}
\end{figure}

\noindent\textbf{Novel View Synthesis.}
As shown in Tab.~\ref{tab:model_comparison_view} and Fig.~\ref{fig:novel-view_table}, HandSCS shows the highest rendering quality under novel viewpoints. Averaged over three subjects, HandSCS improves PSNR by +3.78 dB and SSIM by +0.017, and reduces LPIPS by 0.017 (37\% relative reduction) compared with the strongest baseline GauHuman~\cite{gauhuman}.
Our method preserves fine appearance details such as skin folds, fingernails, and finger boundaries, and introduces fewer artifacts in challenging regions, including occlusions and self-shadowed areas.
LiveHand\cite{livehand} obtains slightly higher SSIM on two subjects. This is expected, as NeRF-based methods interpolate smoothly across sparse viewpoints, while Gaussian-based models may exhibit mild color inconsistencies under difficult viewing angles. Nevertheless, HandSCS consistently produces sharper details, higher-fidelity textures, and significantly fewer rendering artifacts across views. Additional qualitative comparisons are provided in the supplementary material.

\subsection{Ablation Study}
We conduct ablation experiments on the InterHand2.6M dataset using subject test/Capture0 to assess the contribution of each component in HandSCS, including the structural coordinate (Sec.~\ref{sec:3.1}), intra-pose consistency (Sec.~\ref{sec:3.1.3}), and inter-pose consistency (Sec.~\ref{sec:3.2}).

\begin{figure*}[h!]
\vspace{-1em}
    \centering
    \includegraphics[width=\linewidth]{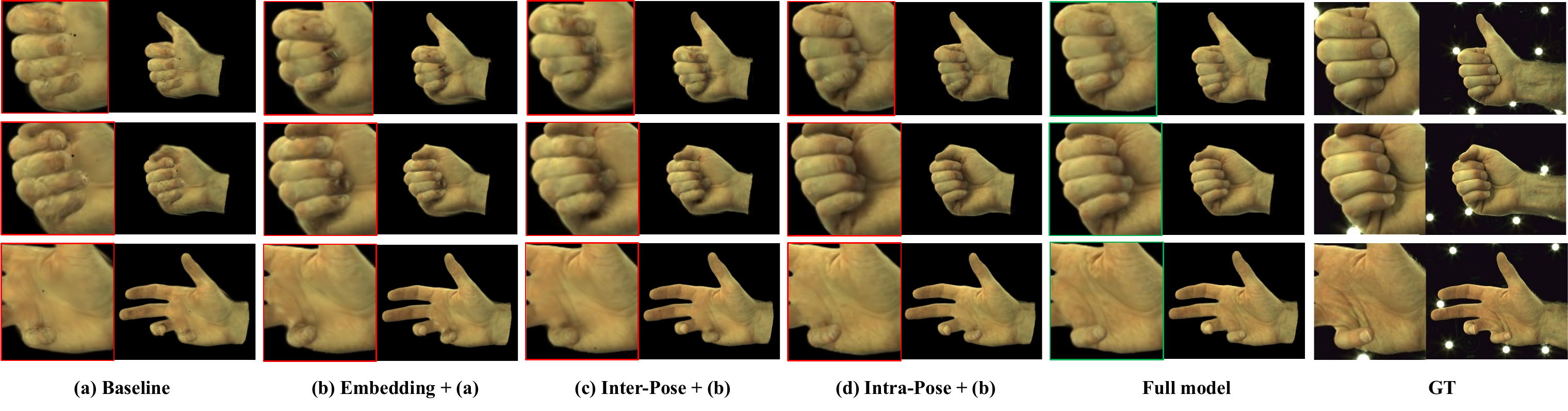}
    \caption{\textbf{Visual comparison of component ablations.}
    The full model reduces artifacts in challenging deformation and self-contact regions, producing clearer boundaries and more consistent details.}
    \label{fig:ablation}
    \vspace{-2em}
\end{figure*}

\noindent{\bf Component Ablation.}
As shown in Tab.~\ref{tab:component_table}, intra-pose consistency via per-Gaussian embeddings (b) improves over the baseline by providing each Gaussian with a learnable identity for non-rigid deformation. 
Inter-pose consistency (c) further enhances global coherence by aligning attributes across structurally corresponding Gaussians. 
The SCS coordinate (d) yields the largest single gain by grounding each Gaussian with an explicit structural representation. 
The full model, combining all components, achieves the best overall performance.
Visually, as shown in Fig.~\ref{fig:ablation}, the baseline shows noise, blurred boundaries, and strong artifacts in contact regions. Embeddings improve stability but still fail under complex deformations and close contact. The consistency loss sharpens global structure, while the SCS coordinate maintains structure consistency such as wrinkles and fingernails. The full model yields the cleanest results with clear boundaries and minimal artifacts.

\begin{figure}[h!]
\vspace{-1em}
\begin{minipage}{0.48\linewidth}
\centering
\resizebox{\linewidth}{!}{
\begin{tabular}{l|ccc}
    \toprule
     Models & \textbf{PSNR} $\uparrow$ & \textbf{SSIM} $\uparrow$ & \textbf{LPIPS} $\downarrow$  \\
    \midrule
    w/o Intra-pose & 32.42 & 0.9636 & 0.0335 \\
    \midrule
    w/o Static Bones & 32.64 & 0.965 & 0.0302 \\
    w/o Dynamic Bones & 32.70 & 0.965 & 0.0311 \\
    \midrule
    Full model & \textbf{32.91} & \textbf{0.966} & \textbf{0.0291} \\
    \bottomrule
\end{tabular}}
\captionof{table}{{\bf Coordinate basis ablation.}
Virtual bones reduce artifacts in high curvature deformed regions but may introduce pose inaccuracies,
while static bones provide a stable structural prior but are less expressive.
Combining both provides the most robust reconstruction.
    }
    \label{tab:coord_basis_table}
\end{minipage}
\hfill
\centering
\begin{minipage}{0.48\linewidth}
\centering
\includegraphics[width=\linewidth]{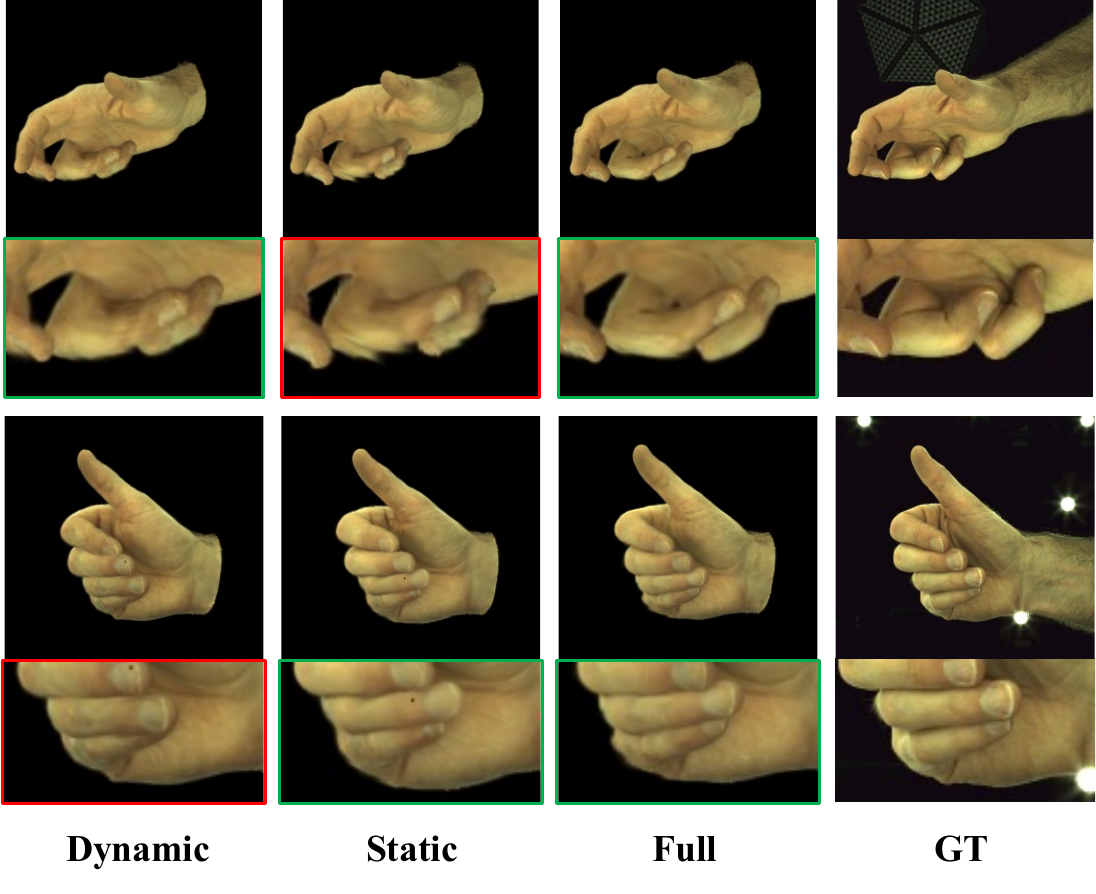}
\caption{\bf{Visual comparison of Coordinate Basis.}}
\label{fig:coord_basis_vis}
\end{minipage}
\vspace{-1em}
\end{figure}

\noindent{\bf{Coordinate Basis.}}
To understand the role of the coordinate basis, we analyze static and virtual bones individually. 
Virtual bones provide complementary structural cues beyond the physical skeleton, suppressing artifacts such as surface glitches and broken boundaries around highly bent fingers, but may occasionally introduce pose inconsistencies due to the lack of strict kinematic constraints. 
Static bones, based on the MANO topology, maintain globally coherent structure and correct overall pose, but their limited diversity makes them less effective in highly deformed regions. 
As shown in Tab.~\ref{tab:coord_basis_table}, removing either component degrades quantitative performance. 
The visual comparison in Fig.~\ref{fig:coord_basis_vis} further shows that combining both provides stable global structure from static bones and expressive local cues from virtual bones, leading to the most robust reconstruction across both rigid and highly articulated poses.

\noindent{\bf Choice of Space for Virtual Bones.}
We compare constructing virtual bones in canonical space and posed space. 
As shown in Tab.~\ref{tab:canonical_posed_space}, canonical space achieves better performance, particularly in LPIPS. 
This is because static bones already operate in the deformed space, capturing each Gaussian's relationship to the posed skeleton, making virtual bones in the same space partially redundant. 
In contrast, canonical-space virtual bones encode the displacement of each Gaussian from its rest-pose configuration, providing complementary structural cues.

\noindent{\bf Virtual Bone Component Ablation.}
We analyze the two factors determining the endpoint positions of virtual bones.
Removing the interpolation parameter $t$ degrades performance, as $t$ enables smooth positioning along bone segments.
Removing the offset term $\Delta$ reduces local deformation flexibility and detail fidelity.
As shown in Tab.~\ref{tab:ablation_of_v_bone}, using both provides complementary benefits and yields the best reconstruction results.

\begin{table}[h!]
\vspace{-1em}
\centering
\scriptsize
\setlength{\tabcolsep}{4pt}

\begin{minipage}[t]{0.48\linewidth}
\centering
\begin{tabular}{lccc}
\toprule
Models & PSNR $\uparrow$ & SSIM $\uparrow$ & LPIPS $\downarrow$ \\
\midrule
Posed Space  & 32.87 & 0.9657 & 0.0303 \\
Canonical Space & 32.91 & 0.9663 & 0.0291 \\
\bottomrule
\end{tabular}

\vspace{4pt}
\caption{Canonical vs. posed space for virtual bones.}
\label{tab:canonical_posed_space}
\end{minipage}
\hfill
\begin{minipage}[t]{0.48\linewidth}
\centering
\begin{tabular}{lccc}
\toprule
Models & PSNR $\uparrow$ & SSIM $\uparrow$ & LPIPS $\downarrow$ \\
\midrule
w/o Intra-pose & 32.42 & 0.9636 & 0.0335 \\
w/o $t$        & 32.67 & 0.9656 & 0.0300 \\
w/o $\Delta$   & 32.69 & 0.9657 & 0.0300 \\
\midrule
Full model & \textbf{32.91} & \textbf{0.9663} & \textbf{0.0291} \\
\bottomrule
\end{tabular}

\vspace{4pt}
\caption{Ablation of virtual bone components.}
\label{tab:ablation_of_v_bone}
\end{minipage}
\vspace{-3em}
\end{table}

\begin{figure}[h!]
    \centering
    \vspace{-2em}
    \includegraphics[width=0.8\linewidth]{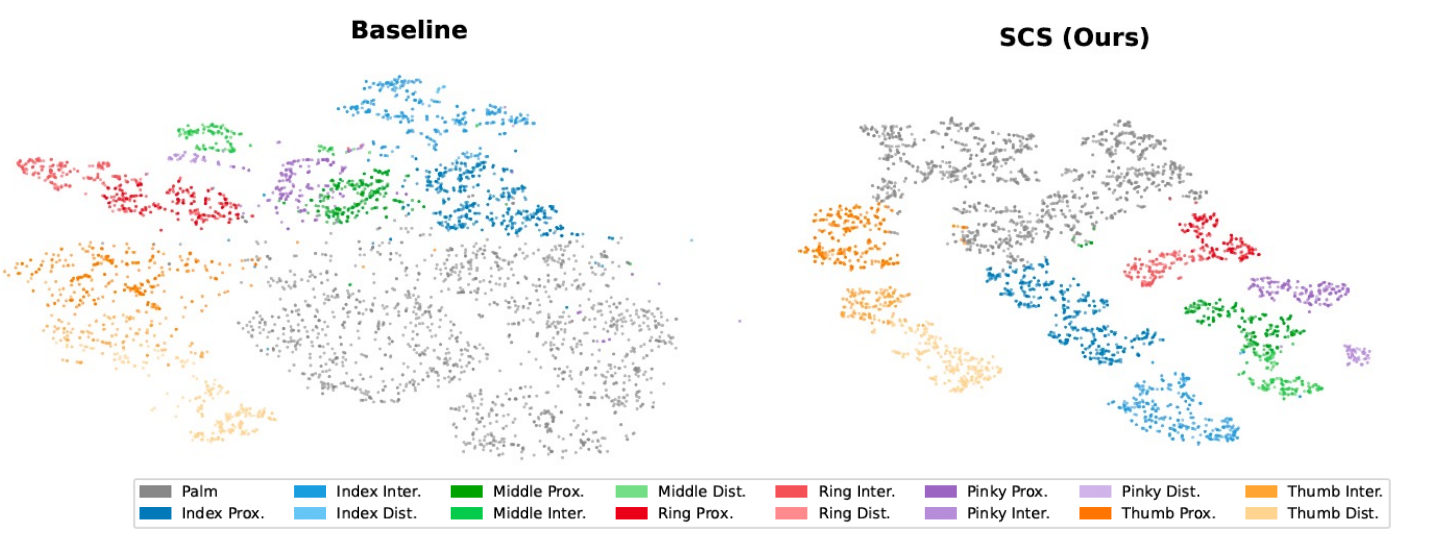}
    \caption{\textbf{t-SNE visualization.}
Without SCS, per-Gaussian features show weak structural separation, while adding SCS leads to clearer clustering by anatomical region (phalange label from LBS argmax).}
    \label{fig:tsne_vis}
    \vspace{-3em}
\end{figure}

\noindent{\bf Visualization of Deformation Features.}
To qualitatively assess the effect of SCS, we visualize the learned per-Gaussian deformation descriptors used by our non-rigid deformation MLP.
Specifically, we embed these descriptors with t-SNE~\cite{tsne} (Fig.~\ref{fig:tsne_vis}) and color each point by its phalange label (LBS weight argmax).
Compared with the baseline without SCS, descriptors from the same anatomical region form more compact clusters, indicating that SCS injects meaningful structural cues into the deformation representation.

\section{Conclusions}
We presented HandSCS, a structure-aware framework for animatable hand avatars based on 3D Gaussian Splatting. 
Our method introduces the Structural Coordinate Space (SCS), which associates each Gaussian primitive with an explicit structural coordinate relative to the articulated hand skeleton. 
By formulating non-rigid deformation and cross-pose consistency within this structural representation, the model better preserves fine geometric details and maintains stable appearance under complex articulations. 
Extensive experiments on the InterHand2.6M dataset demonstrate that HandSCS consistently outperforms existing 3DGS-based and NeRF-based approaches in both quantitative metrics and visual quality. 
Ablation studies further confirm the effectiveness of the structural coordinate and its hybrid static–virtual bone basis, highlighting the importance of explicit structural representations for Gaussian-based avatar modeling and their potential for broader articulated avatars.

\section{Acknowledgement}
This work was supported by the Engineering and Physical Sciences Research Council [grant number EP/Y009800/1], through funding from Responsible Ai UK (KP0016). This research utilised Queen Mary's Apocrita HPC facility, supported by QMUL Research-IT.

\bibliographystyle{splncs04}
\bibliography{main}
\end{document}